\documentclass[11pt,a4paper]{article}
\usepackage[breaklinks=true]{hyperref}
\usepackage[hyperref]{acl2017}
\usepackage{times}
\usepackage{latexsym}

\usepackage{amsmath}
\usepackage{amssymb}
\usepackage{color}
\usepackage{comment}
\usepackage{subcaption}
\usepackage{url}

\usepackage{tikz}
\usepackage{tikz-qtree}
\usetikzlibrary{arrows}
\usetikzlibrary{calc}
\usetikzlibrary{decorations.pathreplacing}
\usetikzlibrary{positioning}

\DeclareMathOperator*{\argmax}{\arg\!\max}
\newcommand{\forwardVector}{\mathbf{f}}
\newcommand{\backwardVector}{\mathbf{b}}
\newcommand{\spanVector}{\mathbf{s}}
\newcommand{\hiddenWeight}{\mathbf{W}}
\newcommand{\hiddenBias}{\mathbf{b}}
\newcommand{\finalWeight}{\mathbf{V}}
\newcommand{\finalVector}{\mathbf{v}}
\newcommand{\labelScore}{s_\mathrm{label}}
\newcommand{\labelScores}{s_\mathrm{labels}}
\newcommand{\spanScore}{s_\mathrm{span}}
\newcommand{\splitScore}{s_\mathrm{split}}
\newcommand{\chartSplitScore}{\tilde{s}_\mathrm{split}}
\newcommand{\bestScore}{s_\mathrm{best}}
\newcommand{\treeScore}{s_\mathrm{tree}}
\newcommand{\topLabelScore}{s_\mathrm{top}}
\newcommand{\middleLabelScore}{s_\mathrm{middle}}
\newcommand{\bottomLabelScore}{s_\mathrm{bottom}}
\newcommand{\leftScore}{s_\mathrm{left}}
\newcommand{\rightScore}{s_\mathrm{right}}

\newcommand{\leftVector}{\mathbf{v}_\mathrm{left}}
\newcommand{\rightVector}{\mathbf{v}_\mathrm{right}}
\newcommand{\leftFeedforward}{f_\mathrm{left}}
\newcommand{\rightFeedforward}{f_\mathrm{right}}
\newcommand{\hiddenVector}{\mathbf{h}}

\aclfinalcopy 

\title{A Minimal Span-Based Neural Constituency Parser}

\author{
  Mitchell Stern \qquad Jacob Andreas \qquad Dan Klein \\
  Computer Science Division \\
  University of California, Berkeley \\
  {\tt \{mitchell,jda,klein\}@cs.berkeley.edu}
}

\date{}

\begin{document}

\maketitle

\begin{abstract}
In this work, we present a minimal neural model for constituency parsing based on independent scoring of labels and spans. We show that this model is not only compatible with classical dynamic programming techniques, but also admits a novel greedy top-down inference algorithm based on recursive partitioning of the input. We demonstrate empirically that both prediction schemes are competitive with recent work, and when combined with basic extensions to the scoring model are capable of achieving state-of-the-art single-model performance on the Penn Treebank (91.79~F1) and strong performance on the French Treebank (82.23~F1).
\end{abstract}

\section{Introduction}

This paper presents a minimal but surprisingly effective span-based neural model for constituency parsing. Recent years have seen a great deal of interest in parsing architectures that make use of recurrent neural network (RNN) representations of input sentences \citep{Vinyals15Parser}. Despite evidence that linear RNN decoders are implicitly able to respect some nontrivial well-formedness constraints on structured outputs \citep{Graves13Seqs}, researchers have consistently found that the best performance is achieved by systems that explicitly require the decoder to generate well-formed tree structures \citep{Chen14Parser}.

There are two general approaches to ensuring this structural consistency. The most common is to encode the output as a sequence of operations within a transition system which constructs trees incrementally. This transforms the parsing problem back into a sequence-to-sequence problem, while making it easy to force the decoder to take only actions guaranteed to produce well-formed outputs. However, transition-based models do not admit fast dynamic programs
and require careful feature engineering to support exact search-based
inference \cite{Le15ShiftReduce}. Moreover, models with recurrent state require complex training procedures to benefit from anything other than greedy decoding \citep{Wiseman16Global}.

An alternative line of work focuses on \emph{chart parsers}, which use log-linear or neural scoring potentials to parameterize a tree-structured dynamic program for maximization or marginalization \citep{Finkel08CRFParser,Durrett15NeuralCRF}. These models enjoy a number of appealing formal properties, including support for exact inference and structured loss functions. However, previous chart-based approaches have required considerable scaffolding beyond a simple well-formedness potential, e.g.\ pre-specification of a complete context-free grammar for generating output structures and  initial pruning of the output space with a weaker model \citep{Hall14CRFParser}. Additionally, we are unaware of any recent chart-based models that achieve results competitive with the best transition-based models.

In this work, we present an extremely simple chart-based neural parser based on independent scoring of labels and spans, and show how this model can be adapted to support a greedy top-down decoding procedure. Our goal is to preserve the basic algorithmic properties of span-oriented (rather than transition-oriented) parse representations, while exploring the extent to which neural representational machinery can replace the additional structure required by existing chart parsers. On the Penn Treebank, our approach outperforms a number of recent models for chart-based and transition-based parsing---including the state-of-the-art models of \citet{Cross16Span} and \citet{Liu16ShiftReduce}---achieving an F1 score of 91.79. We additionally obtain a strong F1 score of 82.23 on the French Treebank.

\section{Model}
\label{sec:model}

A constituency tree can be regarded as a collection of labeled spans over a sentence. Taking this view as a guiding principle, we propose a model with two components, one which assigns scores to span labels and one which assigns scores directly to span existence. The former is used to determine the labeling of the output, and the latter provides its structure.

At the core of both of these components is the issue of span representation. Given that a span's correct label and its quality as a constituent depend heavily on the context in which it appears, we naturally turn to recurrent neural networks as a starting point, since they have previously been shown to capture contextual information suitable for use in a variety of natural language applications \citep{Bahdanau14Neural,Wang15Unified}

In particular, we run a bidirectional LSTM over the input to obtain context-sensitive forward and backward encodings for each position $i$, denoted by $\forwardVector_i$ and $\backwardVector_i$, respectively. Our representation of the span $(i, j)$ is then the concatenatation the vector differences $\forwardVector_j - \forwardVector_i$ and $\backwardVector_i - \backwardVector_j$. This corresponds to a bidirectional version of the LSTM-Minus features first proposed by \citet{Wang16Graph}.

On top of this base, our label and span scoring functions are implemented as one-layer feedforward networks, taking as input the concatenated span difference and producing as output either a vector of label scores or a single span score. More formally, letting $\spanVector_{ij}$ denote the vector representation of span $(i, j)$, we define
\begin{align*}
\labelScores(i, j) &= \finalWeight_\ell g(\hiddenWeight_\ell \spanVector_{ij} + \hiddenBias_\ell), \\
\spanScore(i, j) &= \finalVector_s^\top g(\hiddenWeight_s \spanVector_{ij} + \hiddenBias_s) ,
\end{align*}
where $g$ denotes an elementwise nonlinearity. For notational convenience, we also let the score of an individual label $\ell$ be denoted by
\begin{align*}
\labelScore(i, j, \ell) = [ \labelScores(i, j) ]_\ell ,
\end{align*}
where the right-hand side is the corresponding element of the label score vector.

One potential issue is the existence of unary chains, corresponding to nested labeled spans with the same endpoints. We take the common approach of treating these as additional atomic labels alongside all elementary nonterminals. To accommodate $n$-ary trees, our inventory additionally includes a special empty label $\varnothing$ used for spans that are not themselves full constituents but arise during the course of implicit binarization.

Our model shares several features in common with that of \citet{Cross16Span}. In particular, our representation of spans and the form of our label scoring function were directly inspired by their work, as were our handling of unary chains and our use of an empty label. However, our approach differs in its treatment of structural decisions, and consequently, the inference algorithms we describe below diverge significantly from their transition-based framework.

\section{Chart Parsing}
\label{sec:chart-parsing}

Our basic model is compatible with traditional chart-based dynamic programming. Representing a constituency tree $T$ by its labeled spans,
\begin{align*}
T := \{ (\ell_t, (i_t, j_t)) : t = 1, \dots, |T| \} ,
\end{align*}
we define the score of a tree to be the sum of its constituent label and span scores,
\begin{align*}
\treeScore(T) = \sum_{(\ell, (i, j)) \in T} \left[ \labelScore(i, j, \ell) + \spanScore(i, j) \right] .
\end{align*}
To find the tree with the highest score for a given sentence, we use a modified CKY recursion. As with classical chart parsing, the running time of our procedure is $O(n^3)$ for a sentence of length $n$.

\subsection{Dynamic Program for Inference}
\label{subsec:dynamic-program}

The base case is a span $(i, i+1)$ consisting of a single word. Since every valid tree must include all singleton spans, possibly with empty labels, we need not consider the span score in this case and perform only a single maximization over the choice of label:
\begin{align*}
\bestScore(i, i+1) = \max_\ell \left[ \labelScore(i, i+1, \ell) \right] .
\end{align*}

For a general span $(i, j)$, we define the score of the split $(i, k, j)$ as the sum of its subspan scores,
\begin{align}\label{eq:split-score}
\splitScore(i, k, j) = \spanScore(i, k) + \spanScore(k, j) .
\end{align}
For convenience, we also define an augmented split score incorporating the scores of the corresponding subtrees,
\begin{align*}
\chartSplitScore(i, k, j) & = \splitScore(i, k, j) \\
& \qquad + \bestScore(i, k) + \bestScore(k, j) .
\end{align*}
Using these quantities, we can then write the general joint label and split decision as
\begin{align}\label{eq:best-score-general}
\bestScore(i, j) = \max_{\ell, k} \left[ \labelScore(i, j, \ell) + \chartSplitScore(i, k, j) \right] .
\end{align}
Because our model assigns independent scores to labels and spans, this maximization decomposes into two disjoint subproblems, greatly reducing the size of the state space:
\begin{align*}
\bestScore(i, j) & = \max_\ell \left[ \labelScore(i, j, \ell) \right] \\
& + \max_k \left[ \chartSplitScore(i, k, j) \right] .
\end{align*}

We also note that the span scores $\spanScore(i, j)$ for each span $(i, j)$ in the sentence can be computed once at the beginning of the procedure and shared across different subproblems with common left or right endpoints, allowing for a quadratic rather than cubic number of span score computations.

\subsection{Margin Training}
\label{subsec:chart-margin-training}

Training the model under this inference scheme is accomplished using a margin-based approach. When presented with an example sentence and its corresponding parse tree $T^\ast$, we compute the best prediction under the current model using the above dynamic program,
\begin{align*}
\widehat{T} = \argmax_T \left[ \treeScore(T) \right] .
\end{align*}
If $\widehat{T} = T^\ast$, then our prediction was correct and no changes need to be made. Otherwise, we incur a hinge penalty of the form
\begin{align*}
\max \left( 0,\ 1 - \treeScore(T^\ast) + \treeScore(\widehat{T}) \right)
\end{align*}
to encourage the model to keep a margin of at least $1$ between the gold tree and the best alternative. The loss to be minimized is then the sum of penalties across all training examples.

Prior work has found that it can be beneficial in a variety of applications to incorporate a structured loss function into this margin objective, replacing the hinge penalty above with one of the form
\begin{equation*}
\max\left( 0,\ \Delta(\widehat{T}, T^\ast) - \treeScore(T^\ast) + \treeScore(\widehat{T}) \right)
\end{equation*}
for a loss function $\Delta$ that measures the similarity between the prediction $\widehat{T}$ and the reference $T^\ast$. Here we take $\Delta$ to be a Hamming loss on labeled spans. To incorporate this loss into the training objective, we modify the dynamic program of Section~\ref{subsec:dynamic-program} to support loss-augmented decoding \cite{Taskar05Learning}. Since the label decisions are isolated from the structural decisions, it suffices to replace every occurrence of the label scoring function $\labelScore(i, j, \ell)$ by
\begin{align*}
\labelScore(i, j, \ell) + \mathbf{1}(\ell \ne \ell_{ij}^\ast) ,
\end{align*}
where $\ell_{ij}^\ast$ is the label of span $(i, j)$ in the gold tree $T^\ast$. This has the effect of requiring larger margins between the gold tree and predictions that contain more mistakes, offering a greater degree of robustness and better generalization.

\section{Top-Down Parsing}
\label{sec:top-down-parsing}

\begin{figure*}[t]
\centering

\begin{tikzpicture}[xscale=1.6, baseline=(tennis.base)]
\node[align=center,anchor=base] at (0,0) {PRP \\ She};
\node[align=center,anchor=base] at (1,0) {VBZ \\ enjoys};
\node[align=center,anchor=base] (playing) at (2,0) {VBG \\ playing};
\node[align=center,anchor=base] (tennis) at (3,0) {NN \\ tennis};
\node[align=center,anchor=base] at (4,0) {. \\ .};
\draw[decorate, decoration={brace, raise=-5pt, amplitude=5pt, mirror}, yshift=-5pt] (-1,1) -- (-1,-0.5) node [midway, xshift=-15pt] {input};
\begin{scope}[shift={(-0.5,-15pt)}]
\small
\foreach \i in {0,1,2,3,4,5} \node at (\i,0) {\i};
\end{scope}
\begin{scope}[shift={(0,6.5)}]
\path[{|-|}, shorten <= -20pt, shorten >= -20pt]
(0,-1) edge node[above] (S) {S} (4,-1)
(0,-2) edge node[above] {NP} (0.001,-2)
(1,-2) edge node[above] {$\varnothing$} (4,-2)
(1,-3) edge node[above] {VP} (3,-3)
(4,-3) edge node[above] {$\varnothing$} (4.001,-3)
(1,-4) edge node[above] {$\varnothing$} (1.001,-4)
(2,-4) edge node[above] {S--VP} (3,-4)
(2,-5) edge node[above] {$\varnothing$} (2.001,-5)
(3,-5) edge node[above] {NP} (3.001,-5)
;
\begin{scope}[shift={(0.5,0pt)}]
\path[dashed, shorten <= -0pt, shorten >= -10pt]
(0,-1) edge (0,-1.001)
(3,-2) edge (3,-2.001)
(1,-3) edge (1,-3.001)
(2,-4) edge (2,-4.001)
;
\end{scope}
\draw[->, >=triangle 45] (-1,-1) -- node[sloped, below] {top-down parsing} (-1,-5);
\end{scope}
\node[anchor=base] at ($(playing.base) + (0,-1.5)$) {(a) Execution of the top-down parsing algorithm.};
\end{tikzpicture}
\hspace{10pt}
\begin{tikzpicture}[baseline=(tennis.base), level distance=30pt]
\tikzset{every tree node/.style={align=center,anchor=north}}
\Tree [. S [. NP {PRP \\ She} ] [. VP {VBZ \\ enjoys} [. S [. VP {VBG \\ playing} [. NP \node(tennis){NN \\ tennis}; ] ] ] ] {. \\ .} ]
\node[anchor=base] at ($(tennis.base) + (-1.5, -1.5)$) {(b) Output parse tree.};
\end{tikzpicture}

\caption{An execution of our top-down parsing algorithm (a) and the resulting parse tree (b) for the sentence ``She enjoys playing tennis.'' Part-of-speech tags, shown here together with the words, are predicted externally and are included as part of the input to our system. Beginning with the full sentence span $(0, 5)$, the label S and the split point 1 are predicted, and recursive calls are made on the child spans $(0, 1)$ and $(1, 5)$. The left child span $(0, 1)$ is assigned the label NP, and with no further splits to make, recursion terminates on this branch. The right child span $(1, 5)$ is assigned the empty label $\varnothing$, indicating that it does not represent a constituent in the tree. A split point of 4 is selected, and further recursive calls are made on the grandchild spans $(1, 4)$ and $(4, 5)$. This process of labeling and splitting continues until every branch of recursion bottoms out in singleton spans, at which point the full parse tree can be returned. Note that the unary chain S--VP is produced in a single labeling step.}
\label{fig:top-down-example}
\end{figure*}
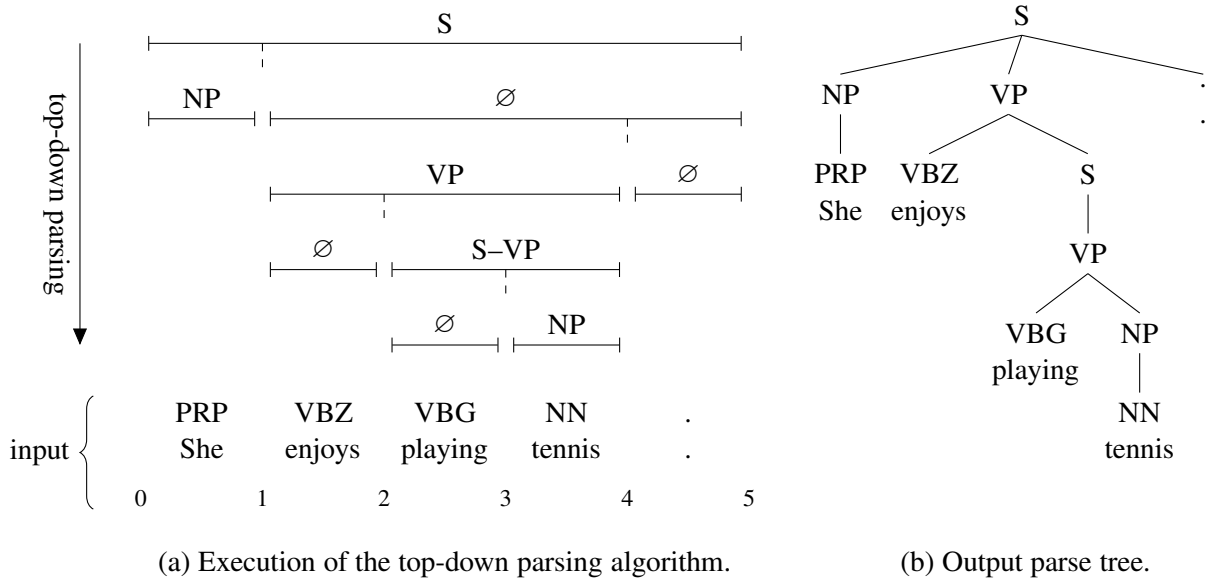

While we have so far motivated our model from the perspective of 
classical chart parsing, it also allows for a novel inference algorithm in which trees are constructed greedily from the top down. At a high level, given a span, we independently assign it a label and pick a split point, then repeat this process for the left and right subspans; the recursion bottoms out with length-one spans that can no longer be split. Figure~\ref{fig:top-down-example} gives an illustration of the process, which we describe in more detail below.

The base case is again a singleton span $(i, i + 1)$, and follows the same form as the base case for the chart parser. In particular, we select the label $\widehat{\ell}$ that satisfies
\begin{align*}
\widehat{\ell} = \argmax_\ell \left[ \labelScore(i, i + 1, \ell) \right] ,
\end{align*}
omitting span scores from consideration since singleton spans cannot be split.

To construct a tree over a general span $(i, j)$, we aim to solve the maximization problem
\begin{align*}
(\widehat{\ell}, \widehat{k}) = \argmax_{\ell, k} \left[ \labelScore(i, j, \ell) + \splitScore(i, k, j) \right] ,
\end{align*}
where $\splitScore(i, k, j)$ is defined as in Equation~\eqref{eq:split-score}. The independence of our label and span scoring functions again yields the decomposed form
\begin{align}\label{eq:top-down-label-split-predictions}
\begin{split}
\widehat{\ell} &= \argmax_\ell \left[ \labelScore(i, j, \ell) \right] , \\
\widehat{k} &= \argmax_k \left[ \splitScore(i, k, j) \right] ,
\end{split}
\end{align}
leading to a significant reduction in the size of the state space.

To generate a tree for the whole sentence, we call this procedure on the full sentence span $(0, n)$ and return the result. As there are $O(n)$ spans each requiring one label evaluation and at most $n - 1$ split point evaluations, the running time of the procedure is $O(n^2)$.

The algorithm outlined here bears a strong resemblance to the chart parsing dynamic program discussed in Section~\ref{sec:chart-parsing}, but differs in one key aspect. When performing inference from the bottom up, we have already computed the scores of all of the subtrees below the current span, and we can take this knowledge into consideration when selecting a split point. In contrast, when producing a tree from the top down, we can only select a split point based on top-level evaluations of span quality, without knowing anything about the subtrees that will be generated below them. This difference is manifested in the augmented split score $\chartSplitScore$ used in the definition of $\bestScore$ in Equation~\eqref{eq:best-score-general}, where the scores of the subtrees associated with a split point are included in the chart recursion but necessarily excluded from the top-down recursion.

While this apparent deficiency may be a cause for concern, we demonstrate the surprising empirical result in Section~\ref{sec:experiments} that there is no loss in performance when moving from the globally-optimal chart parser to the greedy top-down procedure.

\subsection{Margin Training}
\label{subsec:top-down-margin-training}

As with the chart parsing formulation, we also use a margin-based method for learning under the top-down model. However, rather than requiring separation between the scores of full trees, we instead enforce a local margin at every decision point.

For a span $(i, j)$ occurring in the gold tree, let $\ell^\ast$ and $k^\ast$ represent the correct label and split point, and let $\widehat{\ell}$ and $\widehat{k}$ be the predictions made by computing the maximizations in Equation~\eqref{eq:top-down-label-split-predictions}. If $\widehat{\ell} \ne \ell^\ast$, meaning the prediction is incorrect, we incur a hinge penalty of the form
\begin{align*}
\max \left( 0, 1 - \labelScore(i, j, \ell^\ast) + \labelScore(i, j, \widehat{\ell}) \right) .
\end{align*}
Similarly, if $\widehat{k} \ne k^\ast$, we incur a hinge penalty of the form
\begin{align*}
\max \left( 0, 1 - \splitScore(i, k^\ast, j) + \splitScore(i, \widehat{k}, j) \right) .
\end{align*}
To obtain the loss for a given training example, we trace out the actions corresponding to the gold tree and accumulate the above penalties over all decision points. As before, the total loss to be minimized is the sum of losses across all training examples.

Loss augmentation is also beneficial for the local decisions made by the top-down model, and can be implemented in a manner akin to the one discussed in Section~\ref{subsec:chart-margin-training}.

\subsection{Training with Exploration}

The hinge penalties given above are only defined for spans $(i, j)$ that appear in the example tree. The model must therefore be constrained at training time to follow decisions that exactly reproduce the gold tree, since supervision cannot be provided otherwise. As a result, the model is never exposed to its mistakes, which can lead to a lack of calibration and poor performance at test time.

To circumvent this issue, a \textit{dynamic oracle} can be defined to inform the model about correct behavior even after it has deviated from the gold tree. \citet{Cross16Span} propose such an oracle for a related transition-based parsing system, and prove its optimality for the F1 metric on labeled spans. We adapt their result here to obtain a dynamic oracle for the present model with similar guarantees.

The oracle for labeling decisions carries over without modification: the correct label for a span is the label assigned to that span if it is part of the gold tree, or the empty label $\varnothing$ otherwise.

For split point decisions, the oracle can be broken down into two cases. If a span $(i, j)$ appears as a constituent in the gold tree $T$, we let $b(i, j)$ denote the collection of its interior boundary points. For example, if the constituent over $(1, 7)$ has children spanning $(1, 3)$, $(3, 6)$, and $(6, 7)$, then we would have the two interior boundary points, $b(1, 7) = \{ 3, 6 \}$. The oracle for a span appearing in the gold tree is then precisely the output of this function. Otherwise, for spans $(i, j)$ not corresponding to gold constituents, we must instead identify the smallest enclosing gold constituent:
\begin{align*}
(i^\ast, j^\ast) = \min \{ (i', j') \in T : i' \le i < j \le j' \} ,
\end{align*}
where the minimum is taken with respect to the partial ordering induced by span length. The output of the oracle is then the set of interior boundary points of this enclosing span that also lie inside the original, $\{ k \in b(i^\ast, j^\ast) : i < k < j \}$. The proof of correctness is similar to the proof in \citet{Cross16Span}; we refer to the Dynamic Oracle section in their paper for a more detailed discussion.

As presented, the dynamic oracle for split point decisions returns a collection of one or more splits rather than a single correct answer. Any of these is a valid choice, with different splits corresponding to different binarizations of the original $n$-ary tree. We choose to use the leftmost split point for consistency in our implementation, but remark that the oracle split with the highest score could also be chosen at training time to allow for additional flexibility.

Having defined the dynamic oracle for our system, we note that training with exploration can be implemented by a single modification to the procedure described in Section~\ref{subsec:top-down-margin-training}. Local penalties are accumulated as before, but instead of tracing out the decisions required to produce the gold tree, we instead follow the decisions predicted by the model. In this way, supervision is provided at states within the prediction procedure that are more likely to arise at test time when greedy inference is performed.

\section{Scoring and Loss Alternatives}

The model presented in Section~\ref{sec:model} is designed to be as simple as possible. However, there are many variations of the label and span scoring functions that could be explored; we discuss some of the options here.

\subsection{Top-Middle-Bottom Label Scoring}
\label{subsec:top-middle-bottom}

Our basic model treats the empty label, elementary nonterminals, and unary chains each as atomic units, obscuring similarities between unary chains and their component nonterminals or between different unary chains with common prefixes or suffixes. To address this lack of structure, we consider an alternative scoring scheme in which labels are predicted in three parts: a top nonterminal, a middle unary chain, and a bottom nonterminal (each of which is possibly empty).\footnote{In more detail, $\varnothing$ decomposes as \mbox{($\varnothing$, $\varnothing$, $\varnothing$)}, $X$ decomposes as \mbox{($X$, $\varnothing$, $\varnothing$)}, \mbox{$X$--$Y$} decomposes as \mbox{($X$, $\varnothing$, $Y$)}, and \mbox{$X$--$Z_1\text{--}\cdots\text{--}Z_k$--$Y$} decomposes as \mbox{($X$, $Z_1\text{--}\cdots\text{--}Z_k$, $Y$)}.} This not only allows for parameter sharing across labels with common subcomponents, but also has the added benefit of allowing the model to produce novel unary chains at test time.

More precisely, we introduce the decomposition
\begin{align*}
\begin{split}
& \labelScore(i, j, (\ell_t, \ell_m, \ell_b)) = \\
& \quad \topLabelScore(i, j, \ell_t) + \middleLabelScore(i, j, \ell_m) + \bottomLabelScore(i, j, \ell_b) ,
\end{split}
\end{align*}
where $\topLabelScore$, $\middleLabelScore$, and $\bottomLabelScore$ are independent one-layer feedforward networks of the same form as $\labelScore$ that output scores for all label tops, label middle chains, and label bottoms encountered in the training corpus, respectively. The best label for a span $(i, j)$ is then computed by solving the maximization problem
\begin{align*}
& \max_{\ell_t, \ell_m, \ell_b} \left[ \labelScore(i, j, (\ell_t, \ell_m, \ell_b)) \right] ,
\end{align*}
which decomposes into three independent subproblems corresponding to the three label components. The final label is obtained by concatenating $\ell_t$, $\ell_m$, and $\ell_b$, with empty components being omitted from the concatenation.

\subsection{Left and Right Span Scoring}
\label{subsec:left-right-scoring}

The basic model uses the same span scoring function $\spanScore$ to assign a score to the left and right subspans of a given span. One simple extension is to replace this by a pair of distinct left and right feedforward networks of the same form, giving the decomposition
\begin{align*}
\splitScore(i, k, j) = \leftScore(i, k) + \rightScore(k, j) .
\end{align*}

\subsection{Span Concatenation Scoring}
\label{subsec:concatenation-scoring}

Since span scores are only used to score splits in our model, we also consider directly scoring a split by feeding the concatenation of the span representations of the left and right subspans through a single feedforward network, giving
\begin{align*}
\splitScore(i, k, j) = \finalVector_s^\top g \left( \hiddenWeight_s [ \spanVector_{ik}; \spanVector_{kj} ] + \hiddenBias_s \right) .
\end{align*}
This is similar to the structural scoring function used by \citet{Cross16Span}, although whereas they additionally include features for the outside spans $(0, i)$ and $(j, n)$ in their concatenation, we omit these from our implementation, finding that they do not improve performance.

\subsection{Deep Biaffine Span Scoring}
\label{subsec:deep-biaffine-scoring}

Inspired by the success of deep biaffine scoring in recent work by \citet{Dozat16Deep} for dependency parsing, we also consider a split scoring function of a similar form for our model. Specifically, we let $\hiddenVector_{ik} = \leftFeedforward(\spanVector_{ik})$ and $\hiddenVector_{kj} = \rightFeedforward(\spanVector_{kj})$ be deep left and right span representations obtained by passing the child vectors through corresponding left and right feedforward networks. We then define the biaffine split scoring function
\begin{align*}
\splitScore(i, k, j) = \hiddenVector_{ik}^\top \hiddenWeight_s \hiddenVector_{kj} + \leftVector^\top \hiddenVector_{ik} + \rightVector^\top \hiddenVector_{kj} ,
\end{align*}
which consists of the sum of a bilinear form between the two hidden representations together with two inner products.

\subsection{Structured Label Loss}
\label{subsec:structured-label-loss}

The three-way label scoring scheme described in Section~\ref{subsec:top-middle-bottom} offers one path towards the incorporation of label structure into the model. We additionally consider a structured Hamming loss on labels. More specifically, given two labels $\ell_1$ and $\ell_2$ consisting of zero or more nonterminals, we define the loss as $| \ell_1 \setminus \ell_2 | + | \ell_2 \setminus \ell_1 |$, treating each label as a multiset of nonterminals. This structured loss can be incorporated into the training process using the methods described in Sections~\ref{subsec:chart-margin-training} and~\ref{subsec:top-down-margin-training}.

\section{Experiments}
\label{sec:experiments}

\begin{table*}[t]
\centering

\begin{subtable}{0.49\linewidth}
\centering
\resizebox{\linewidth}{!}{\begin{tabular}{c|cccc}
\multicolumn{5}{c}{\bf WSJ Dev, Atomic Labels, Basic 0-1 Label Loss} \\ \hline
Parser & Minimal & Left-Right & Concat. & Biaffine \\ \hline
Chart    & 91.95 & 92.09 & 92.15 & 91.96 \\
Top-Down & 92.16 & 92.25 & 92.24 & 92.14 \\
\end{tabular}}
\caption{\label{tab:results-atomic-zero-one}}
\end{subtable}
\hfill
\begin{subtable}{0.49\linewidth}
\centering
\resizebox{\linewidth}{!}{\begin{tabular}{c|cccc}
\multicolumn{5}{c}{\bf WSJ Dev, Atomic Labels, Structured Label Loss} \\ \hline
Parser & Minimal & Left-Right & Concat. & Biaffine \\ \hline
Chart    & 91.86 & 92.12 & 92.09 & 91.95 \\
Top-Down & 92.12 & 92.31 & 92.26 & 92.20 \\
\end{tabular}}
\caption{\label{tab:results-atomic-hamming}}
\end{subtable}

\vspace{1em}

\begin{subtable}{0.49\linewidth}
\centering
\resizebox{\linewidth}{!}{\begin{tabular}{c|cccc}
\multicolumn{5}{c}{\bf WSJ Dev, 3-Part Labels, Basic 0-1 Label Loss} \\ \hline
Parser & Minimal & Left-Right & Concat. & Biaffine \\ \hline
Chart    & 92.08 & 92.05 & 91.94 & 91.79 \\
Top-Down & 92.12 & 92.18 & 92.14 & 92.02 \\
\end{tabular}}
\caption{\label{tab:results-top-middle-bottom-zero-one}}
\end{subtable}
\hfill
\begin{subtable}{0.49\linewidth}
\centering
\resizebox{\linewidth}{!}{\begin{tabular}{c|cccc}
\multicolumn{5}{c}{\bf WSJ Dev, 3-Part Labels, Structured Label Loss} \\ \hline
Parser & Minimal & Left-Right & Concat. & Biaffine \\ \hline
Chart    & 91.92 & 91.96 & 91.97 & 91.78 \\
Top-Down & 91.98 & 92.27 & 92.17 & 92.06 \\
\end{tabular}}
\caption{\label{tab:results-top-middle-bottom-hamming}}
\end{subtable}

\caption{Development F1 scores on the Penn Treebank. Each table corresponds to a particular choice of label loss (either the basic 0-1 loss or the structured Hamming label loss of Section~\ref{subsec:structured-label-loss}) and labeling scheme (either the basic atomic scheme or the top-middle-bottom labeling scheme of Section~\ref{subsec:top-middle-bottom}). The columns within each table correspond to different split scoring schemes: basic minimal scoring, the left-right scoring of Section~\ref{subsec:left-right-scoring}, the concatenation scoring of Section~\ref{subsec:concatenation-scoring}, and the deep biaffine scoring of Section~\ref{subsec:deep-biaffine-scoring}.}

\end{table*}

We first describe the general setup used for our experiments. We use the Penn Treebank~\citep{Marcus93Building} for our English experiments, with standard splits of sections 2-21 for training, section 22 for development, and section 23 for testing. We use the French Treebank from the SPMRL 2014 shared task~\citep{Seddah14Introducing} with its provided splits for our French experiments. No token preprocessing is performed, and only a single \texttt{<UNK>} token is used for unknown words at test time. The inputs to our system are concatenations of 100-dimensional word embeddings and 50-dimensional part-of-speech embeddings. In the case of the French Treebank, we also include 50-dimensional embeddings of each morphological tag. We use automatically predicted tags for training and testing, obtaining predicted part-of-speech tags for the Penn Treebank using the Stanford tagger~\citep{Toutanova03Tagging} with 10-way jackknifing, and using the provided predicted part-of-speech and morphological tags for the French Treebank.  Words are replaced by \texttt{<UNK>} with probability $1 / (1 + \text{freq}(w))$ during training, where $\text{freq}(w)$ is the frequency of $w$ in the training data.

We use a two-layer bidirectional LSTM for our base span features. Dropout with a ratio selected from \{0.2, 0.3, 0.4\} is applied to all non-recurrent connections of the LSTM, including its inputs and outputs. We tie the hidden dimension of the LSTM and all feedforward networks, selecting a size from \{150, 200, 250\}. All parameters (including word and tag embeddings) are randomly initialized using Glorot initialization \citep{Glorot10Understanding}, and are tuned on development set performance. We use the Adam optimizer~\citep{Kingma14Adam} with its default settings for optimization, with a batch size of 10. Our system is implemented in C++ using the DyNet neural network library \citep{Neubig17Dynet}.

We begin by training the minimal version of our proposed chart and top-down parsers on the Penn Treebank. Out of the box, we obtain test F1 scores of 91.69 for the chart parser and 91.58 for the top-down parser. The higher of these matches the recent state-of-the-art score of 91.7 reported by \citet{Liu16ShiftReduce}, demonstrating that our simple neural parsing system is already capable of achieving strong results.

Building on this, we explore the effects of different split scoring functions when using either the basic 0-1 label loss or the structured label loss discussed in Section~\ref{subsec:structured-label-loss}. Our results are presented in Tables~\ref{tab:results-atomic-zero-one} and~\ref{tab:results-atomic-hamming}.

We observe that regardless of the label loss, the minimal and deep biaffine split scoring schemes perform a notch below the left-right and concatenation scoring schemes. That the minimal scoring scheme performs worse than the left-right scheme is unsurprising, since the latter is a strict generalization of the former. It is evident, however, that joint scoring of left and right subspans is not required for strong results---in fact, the left-right scheme which scores child subspans in isolation slightly outperforms the concatenation scheme in all but one case, and is stronger than the deep biaffine scoring function across the board.

Comparing results across the choice of label loss, however, we find that fewer trends are apparent. The scores obtained by training with a \mbox{0-1} loss are all within 0.1 of those obtained using a structured Hamming loss, being slightly higher in four out of eight cases and slightly lower in the other half. This leads us to conclude that the more elementary approach is sufficient when selecting atomic labels from a fixed inventory.

We also perform the same set of experiments under the setting where the top-middle-bottom label scoring function described in Section~\ref{subsec:top-middle-bottom} is used in place of an atomic label scoring function. These results are shown in Tables~\ref{tab:results-top-middle-bottom-zero-one} and~\ref{tab:results-top-middle-bottom-hamming}.

A priori, we might expect that exposing additional structure would allow the model to make better predictions, but on the whole we find that the scores in this set of experiments are worse than those in the previous set. Trends similar to before hold across the different choices of scoring functions, though in this case the minimal setting has scores closer to those of the left-right setting, even exceeding its performance in the case of a chart parser with a 0-1 label loss.

\begin{table}[t]
\centering
\resizebox{\linewidth}{!}{\begin{tabular}{l|ccc}
\multicolumn{4}{c}{\bf Final Parsing Results on Penn Treebank} \\ \hline
Parser & LR & LP & F1 \\ \hline
\citet{Durrett15NeuralCRF} & -- & -- & 91.1 \\
\citet{Vinyals15Parser} & -- & -- & 88.3 \\
\citet{Dyer16RNNG} & -- & -- & 89.8 \\
\citet{Cross16Span} & 90.5 & 92.1 & 91.3 \\
\citet{Liu16ShiftReduce} & 91.3 & 92.1 & 91.7 \\ \hline
Best Chart Parser & 90.63 & 92.98 & 91.79 \\
Best Top-Down Parser & 90.35 & 93.23 & 91.77 \\ 
\end{tabular}}
\caption{Comparison of final test F1 scores on the Penn Treebank. Here we only include scores from single-model parsers trained without external parse data.}
\label{tab:ptb-test}
\end{table}

Our final test results are given in Table~\ref{tab:ptb-test}, along with the results of other recent single-model parsers trained without external parse data. We achieve a new state-of-the-art F1 score of 91.79 with our best model. Interestingly, we observe that our parsers have a noticeably higher gap between precision and recall than do other top parsers, perhaps in part owing to the structured label loss which penalizes mismatching nonterminals more heavily than it does a nonterminal and empty label mismatch. In addition, there is little difference between the best top-down model and the best chart model, indicating that global normalization is not required to achieve strong results. Processing one sentence at a time on a \texttt{c4.4xlarge} Amazon EC2 instance, our best chart and top-down parsers operate at speeds of 20.3 sentences per second and 75.5 sentences per second, respectively, as measured on the test set.

\begin{table}[t]
\centering
\resizebox{\linewidth}{!}{\begin{tabular}{l|ccc}
\multicolumn{4}{c}{\bf Final Parsing Results on French Treebank} \\ \hline
Parser & LR & LP & F1 \\ \hline
\citet{Bjorkelund14IMS} & -- & -- & 82.53 \\
\citet{Durrett15NeuralCRF} & -- & -- & 81.25 \\
\citet{Cross16Span} & 81.90 & 84.77 & 83.11 \\ \hline
Best Chart Parser & 80.26 & 84.12 & 82.14 \\
Best Top-Down Parser & 79.60 & 85.05 & 82.23 \\
\end{tabular}}
\caption{Comparison of final test F1 scores on the French Treebank.}
\label{tab:french-test}
\end{table}

We additionally train parsers on the French Treebank using the same settings from our English experiments, selecting the best model of each type based on development performance. We list our test results along with those of several other recent papers in Table~\ref{tab:french-test}. Although we fall short of the scores obtained by \citet{Cross16Span}, we achieve competitive performance relative to the neural CRF parser of \citet{Durrett15NeuralCRF}.

\section{Related Work}

Many early successful approaches to constituency parsing focused on rich modeling of correlations in the \emph{output space}, typically by engineering proabilistic context-free grammars with state spaces enriched to capture long-distance dependencies and lexical phenomena \citep{Collins03Parser,Klein03Unlex,Petrov07Parser}. By contrast, the approach we have described here continues a recent line of work on direct modeling of correlations in the \emph{input space}, by using rich feature representations to parameterize local potentials that interact with a comparatively unconstrained structured decoder. As noted in the introduction, this class of feature-based tree scoring functions can be implemented with either a linear transition system \citep{Chen14Parser} or a global decoder \citep{Finkel08CRFParser}. \citet{Kiperwasser16Parser} describe an approach closely related to ours but targeted at dependency formalisms, and which easily accommodates both sparse log-linear scoring models \citep{Hall14CRFParser} and deep neural potentials \citep{Henderson04NeuralParser,Ballesteros16Exploration}.

The best-performing constituency parsers in the last two years have largely been transition-based rather than global; examples include the models of \citet{Dyer16RNNG}, \citet{Cross16Span} and \citet{Liu16ShiftReduce}. The present work takes many of the insights developed in these models (e.g.\ the recurrent representation of spans \citep{Kiperwasser16Parser}, and the use of a dynamic oracle and exploration policy during training \citep{Goldberg13Oracle}) and extends these insights to span-oriented models, which support a wider range of decoding procedures. Our approach differs from other recent chart-based neural models (e.g.\ \citet{Durrett15NeuralCRF}) in the use of a recurrent input representation, structured loss function, and comparatively simple parameterization of the scoring function. In addition to the globally optimal decoding procedures for which these models were designed, and in contrast 
to the left-to-right decoder typically employed by transition-based models, our model admits an additional greedy top-to-bottom inference procedure.

\section{Conclusion}

We have presented a minimal span-oriented parser that uses a recurrent input representation to score trees with a sum of independent potentials on their constituent spans and labels. Our model supports both exact chart-based decoding and a novel top-down inference procedure. Both approaches achieve state-of-the-art performance on the Penn Treebank, and our best model achieves competitive performance on the French Treebank. Our experiments show that many of the key insights from recent neural transition-based approaches to parsing can be easily ported to the chart parsing setting, resulting in a pair of extremely simple models that nonetheless achieve excellent performance.

\section*{Acknowledgments}

We would like to thank Nick Altieri and the anonymous reviewers for their valuable comments and suggestions. MS is supported by an NSF Graduate Research Fellowship. JA is supported by a Facebook graduate fellowship and a Berkeley AI / Huawei fellowship.

\bibliography{acl2017,jacob}
\bibliographystyle{acl_natbib}

\end{document}